\documentclass[10pt,twocolumn,letterpaper]{article}

\usepackage[pagenumbers]{cvpr} 

\usepackage{graphicx}
\usepackage{amsmath}
\usepackage{amssymb}
\usepackage{booktabs}
\usepackage{booktabs}
\usepackage{multirow}
\usepackage{amsmath,amssymb}
\usepackage{float}

\usepackage[dvipsnames,table,xcdraw]{xcolor}
\usepackage[accsupp]{axessibility}  
\usepackage{pifont}

\definecolor{myorange}{rgb}{1, 0.647, 0}
\definecolor{myblue}{rgb}{.118, 0.565, 1}
\graphicspath{Figures/}

\usepackage[pagebackref,breaklinks,colorlinks]{hyperref}

\usepackage[capitalize]{cleveref}
\crefname{section}{Sec.}{Secs.}
\Crefname{section}{Section}{Sections}
\Crefname{table}{Table}{Tables}
\crefname{table}{Tab.}{Tabs.}

\def\cvprPaperID{*****} 
\def\confName{CVPR}
\def\confYear{2023}

\begin{document}

\title{Dilated-UNet: A Fast and Accurate Medical Image Segmentation Approach using a Dilated Transformer and U-Net Architecture}


\author{%
    Davoud Saadati,\qquad Omid Nejati Manzari\footnotemark[1], \qquad Sattar Mirzakuchaki  \\
    School of Electrical Engineering, Iran University of Science and Technology,\\ Tehran, Iran \\
 	{d\_saadati@elec.iust.ac.ir}\\
 	{omid\_nejaty@alumni.iust.ac.ir}\\
    {m\_kuchaki@iust.ac.ir}}

\maketitle
\renewcommand{\thefootnote}{\fnsymbol{footnote}}
\footnotetext[1]{Corresponding author.}

\begin{abstract}
Medical image segmentation is crucial for the development of computer-aided diagnostic and therapeutic systems, but still faces numerous difficulties. In recent years, the commonly used encoder-decoder architecture based on CNNs has been applied effectively in medical image segmentation, but has limitations in terms of learning global context and spatial relationships. Some researchers have attempted to incorporate transformers into both the decoder and encoder components, with promising results, but this approach still requires further improvement due to its high computational complexity. This paper introduces Dilated-UNet, which combines a Dilated Transformer block with the U-Net architecture for accurate and fast medical image segmentation. Image patches are transformed into tokens and fed into the U-shaped encoder-decoder architecture, with skip-connections for local-global semantic feature learning. The encoder uses a hierarchical Dilated Transformer with a combination of Neighborhood Attention and Dilated Neighborhood Attention Transformer to extract local and sparse global attention. The results of our experiments show that Dilated-UNet outperforms other models on several challenging medical image segmentation datasets, such as ISIC and Synapse.  Code is available at \url{https://github.com/Omid-Nejati/Dilated_Unet}.
\end{abstract}


\section{Introduction}
Early detection of diseases is vital in the healthcare industry as it can help determine the extent and progression of disorders in their initial stages. Medical image segmentation, which involves the precise and automatic identification of anatomical structures and other significant areas, is a critical component of medical data analysis~\cite{AATSN}. It plays a crucial role in computer-assisted diagnosis~\cite{pulmonary} evaluation, surgical preparation~\cite{MAF}, and treatment planning~\cite{chandran2023memu}. It enables the rapid detection and localization of lesion boundaries, which can facilitate the early identification of tumor and cancerous areas. 
It primarily involves the division of target organs and tissues into shapes and volumes using Classification at the pixel level~\cite{zhu2022pcan}. The process of manually annotating~\cite{QRS} in clinical applications is demanding and taking a lot of time, with a risk of human error. The utilization of this method has the potential to enhance diagnostic procedures, raise the chances of detecting tumors~\cite{ranjbarzadeh2022brain, ranjbarzadeh2022breast}, and improve the efficiency of clinicians, which ultimately leads to better outcomes for patients~\cite{jia2021prediction}. For several years, researchers have prioritized the development of automated medical image segmentation to decrease the manual annotation workload. With the increasing use of deep learning in computer vision, researchers have been encouraged to investigate how it can assist in computer-assisted diagnosis.

For a considerable amount of time, fully convolutional networks (FCNs), as well as convolutional neural networks (CNNs) more broadly, have been the leading approach in deep learning and have been extensively utilized for medical image segmentation tasks. Various architectures built on CNNs have been utilized for image segmentation~\cite{shelhamer2017fully}. However, these architectures face a significant challenge, important details are often lost in the deeper layers of the network. To address this issue medical image segmentation architectures, employ a symmetrical top-down encoder-decoder design, a family of U-shaped networks was developed, inspired by the influential U-Net structure~\cite{ronneberger2015u}. The U-Net architecture, which has been a major advancement in segmentation algorithms, was created by incorporating a horizontal propagation of intermediate signals (skip connections) into the conventional symmetric top-down encoder-decoder structures. This allows the structure to compress the input image into a latent space and subsequently decode the positions of the image's regions of interest. The UNet's success is largely attributed to its fully convolutional design, as it doesn't incorporate any trainable parameters that are non-convolutional based in its structure. After this approach, numerous algorithms have emerged for medical image and volume segmentation, such as the U-Net++~\cite{zhou2018unet}. These architectures employ skip connections to merge high-level features from the encoder path with detailed features from the decoder path.  Despite their achievement in computer vision, CNN models face challenges in understanding global contexts and long-range dependencies due to their weight-sharing and locality nature of Convolution operations. 
The limited receptive field and inductive bias of CNNs can cause inaccurate segment categorization in image segmentation~\cite{dosovitskiy2020image}, leading to reduced performance. To enhance the global context comprehension of CNNs, several methods have been proposed, including the incorporation of attention mechanisms and expanding the size of the convolution kernel by dilating it, thus increasing its receptive field. However, these methods also have their own drawbacks

Transformers have achieved great success in natural language processing (NLP) tasks due to their efficient handling of long-range sequence dependencies~\cite{vaswani2017attention}. Transformer is a deep neural network based primarily on self-attention, extract intrinsic features that enable efficient construction of long-range dependencies and capture of global contexts in sequences of tokens. However, the larger size, noise, and redundant modalities of images make it a more challenging task compared to text. Researchers aim to apply transformers in computer vision tasks because of its powerful representation abilities, such as DETR~\cite{carion2020end} for object detection, SETR~\cite{zheng2021rethinking} for semantic segmentation, Vision Transformer (ViT)~\cite{dosovitskiy2020image} and DeiT~\cite{touvron2021training} for image recognition. ViT outperforms CNNs on benchmark imaging tasks, and recent advancements including LNL~\cite{manzari2023robust}, CCT~\cite{hassani2021escaping}, and CvT~\cite{wu2021cvt} propose efficient locality inductive bias to address the challenge of transformers for computer vision requiring large amounts of data to learn. Variants of Transformers like the Swin Transformer~\cite{liu2021swin} aim to reduce the computational complexity and high data requirements of vision transformers. 

Although vision transformers have the ability to model global context, their self-attention mechanism may cause low-level features to be overlooked. To address this issue, hybrid Transformer-CNN approaches have been proposed, which blend the CNN's ability to capture local features with the transformer's capacity to model long-range dependencies. TransUnet~\cite{chen2021transunet}, LeVit-Unet~\cite{xu2021levit} and MedViT~\cite{manzari2023medvit} utilize both transformers and CNNs to capture both high-level and low-level features in medical image segmentation. Nonetheless, these methods face some barriers that hinder their ability to achieve improved results. Firstly, they struggle to seamlessly blend high-level and low-level features while preserving feature coherence. Secondly, the hierarchical encoder generates valuable multi-scale information, but the architecture does not effectively leverage this information to improve its performance.

The aim of this paper is to enhance hierarchical transformers for more accurate medical image segmentation by presenting the Dilated-Unet. The proposed architecture is based on the Transformer architecture and adopts a U-shaped design that comprises of four key components: the encoder, the bottleneck, the decoder, and the skip connections. All these components are built using the Dilated Transformer block~\cite{hassani2022dilated}.
The input medical images are divided into small image patches, and each patch is treated as a token. These image patches are then fed into the Transformer-based encoder, which learns local representations using the NA mechanism and sparse global representations using the DiNA mechanism. The extracted context features are then increased in resolution by the decoder with a patch-expanding module and combined with multiscale features from the encoder to restore the spatial resolution of the feature maps and perform segmentation prediction.
A skip connection module is introduced. This module transfers decoder features forward into the encoder blocks, thereby encouraging feature reuse and providing localization information.
Experiments on skin segmentation~\cite{codella2019skin} and multi-organ~\cite{synapse2015ct} datasets have shown that the Dilated-Unet has exceptional segmentation accuracy and robust generalization ability. In summary, the key contributions of this work can be outlined as follows:

\begin{itemize}
\item The Dilated-Unet architecture is an innovative approach for image segmentation tasks that brings together the benefits of Dilated Transformer blocks and U-shaped Encoder-Decoder networks with skip connections. 

\item  The encoder utilizes both DiNA's sparse global attention and NA's local attention to capture both detailed and contextual information. The decoder then uses a patch expanding layer to upscale the features to the input resolution for pixel-level predictions without relying on traditional convolution or interpolation operations.

\item  Experiments showed that a gradual change in dilation throughout the model enhances the effectiveness of the Transformer's receptive fields, leading to the creation of the Dilated-Unet. This architecture combines the strengths of both Dilated Transformer blocks and U-shaped Encoder-Decoder networks with skip connections for improved performance in image segmentation tasks.
\end{itemize}

\section{Literature review}

\textbf{Convolutional Networks}. CNNs have become widely accepted as the benchmark for various computer vision tasks~\cite{azad2021stacked}. Particularly excelling in the task of image segmentation, where each pixel is assigned a class label. The innovation in this field is the FCN~\cite{shelhamer2017fully} model structure, which is an encoder-decoder structure. To address the issue of information loss during the encoding process, residual connections were used to combine the encoding process in the FCN model. Other models have also been proposed that combine the outputs of various layers~\cite{ronneberger2015u, badrinarayanan2017segnet, noh2015learning}.

The initial CNN model proposed for medical image segmentation was U-Net~\cite{ronneberger2015u}, which is based on FCNs. Its primary innovation involves merging high-resolution feature maps from the encoder network with the upsampled features during upsampling, leading to improved representation learning and more precise results. After the advent of U-Net, other researchers turned their attention to using U-shaped encoder-decoder architectures. In In Unet++~\cite{zhou2018unet}, a dense skip connection is included between modules, rather than transmitting high-resolution feature maps directly from the encoder to the decoder. Att U-Net~\cite{oktay2018attention} was one of the first works to incorporate attention models in pancreas segmentation by using a gating mechanism in the flow of features from the encoder to the decoder of a U-Net. This allows the network to focus on crucial objects and ignore unimportant regions. 
The FocusNet~\cite{kaul2019focusnet} model utilizes a two-part encoder-decoder structure that incorporates attention gating to transfer important features from the decoder of one UNet to the encoder of the next. The 3D U-Net~\cite{cciccek20163d} was developed to enable volumetric segmentation in three dimensions as an extension of the original U-Net. A highly influential variation of the UNet architecture is nnUNet~\cite{isensee2021nnu}, which can automatically process data and select an appropriate network structure for a given task without requiring any manual intervention.

Different methods have been created to address the limited receptive field issue in CNNs, which results in a lack of context information and causes segmentation failure in complex areas like boundaries. The study in~\cite{yu2015multi} leverages the capability of dilated convolutions to increase the receptive field without reducing resolution or coverage, resulting in the creation of a novel convolutional network module specifically intended for dense prediction. This module incorporates dilated convolutions to aggregate multi-scale contextual information while preserving resolution. A dual attention network was proposed in~\cite{fu2019dual} that models semantic interdependencies using self-attention mechanism in both spatial and channel dimensions. In another study, \cite{zhao2017pyramid} utilized a pyramid pooling-based approach to aggregate global information at various feature scales. Despite modifications to CNNs, long-range semantic dependency remains a challenge for them.

\textbf{Vision Transformers.} The Vision Transformer (ViT)~\cite{dosovitskiy2020image}, a pioneering model, was a significant step towards adapting Transformers for vision tasks and achieved satisfactory results in image classification. ViT intuitively divides the image into multiple partitions (tokens), processes them through a Transformer encoder, and uses an MLP layer for classification. Although this structure appears efficient, it has significant drawbacks. The quadratic computational requirements of Transformers make them unsuitable for dense prediction with high-resolution images. They effectively capture global context and long-range relationships but lack the ability to capture essential low-level pixel information, resulting in inaccurate segmentation. various alternatives to ViT were proposed in the literature, such as the LeViT~\cite{graham2021levit}, Swin Transformer~\cite{liu2021swin}, and Twins~\cite{chu2021twins}. To address the issue of model complexity and high data requirements in regular Transformers, the Swin Transformer proposed a hierarchical Vision Transformer with local self-attention computation using non-overlapping partial windows. This approach splits the image patches into windows and limits self-attention calculations by applying the Transformer only within the patches inside each window.

\begin{figure*}[ht]
	\centering
	\includegraphics[width=0.95\textwidth]{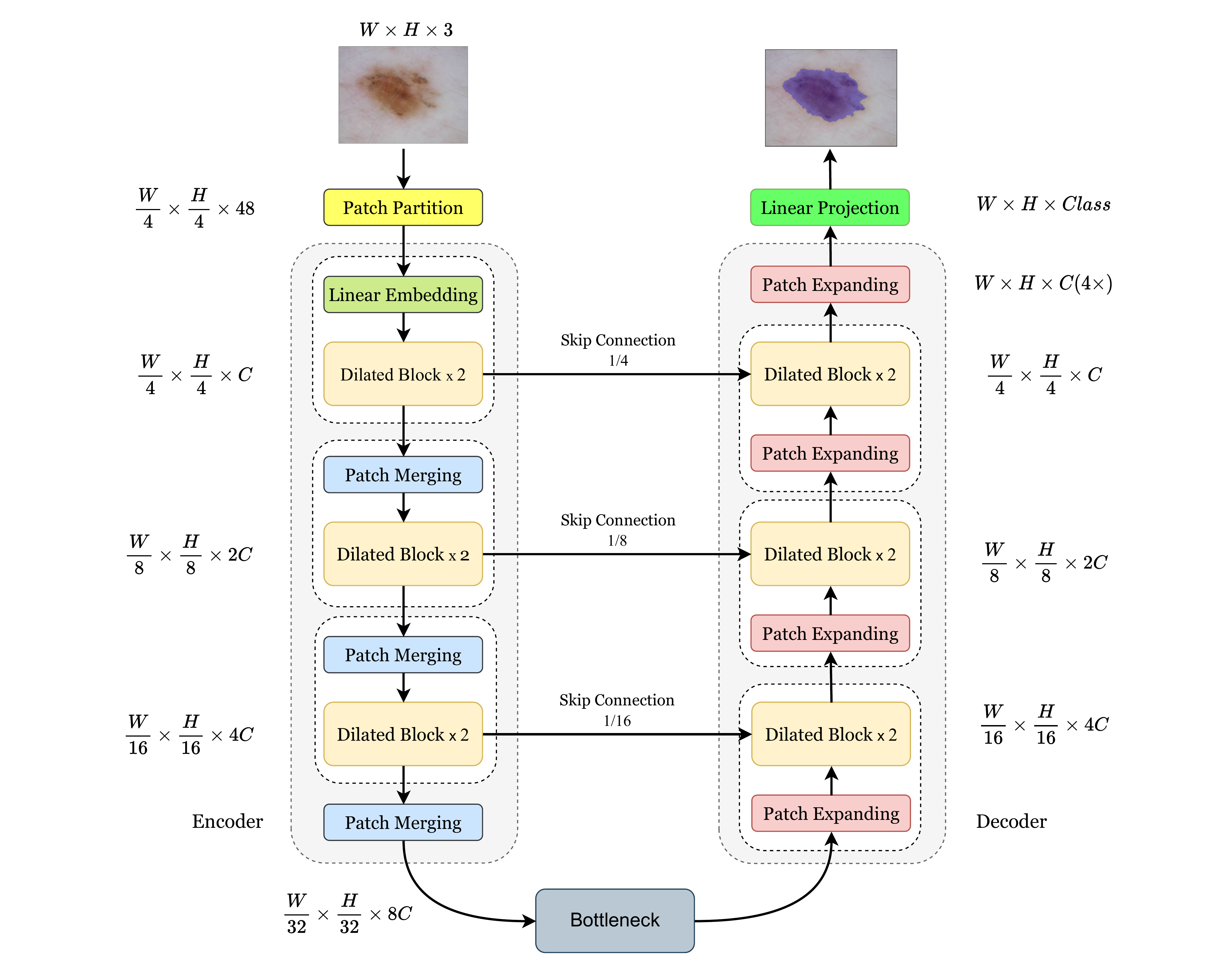}
 \vspace{1em}
	\caption{The overall architecture of Dilated-Unet, which is composed of encoder, decoder, bottleneck, and skip connections. Decoder, encoder, and bottleneck are all built based on Dilated transformer block.}
	\label{fig:overal}
	\vspace{1.5em}
\end{figure*}

\textbf{Hybrid models.} Most existing research in medical image segmentation focuses on creating hybrid models that combine Transformers and CNNs for feature processing. Like the Attention UNet~\cite{oktay2018attention}, the UNet Transformer~\cite{petit2021u} improved CNNs by incorporating multi-head attention within the skip connections. The TransUNet~\cite{chen2021transunet} is one of the pioneering hybrid Transformer-CNN models for medical image segmentation. It combines a transformer encoder and a cascaded convolutional decoder, leveraging the strengths of both CNNs and transformers to extract both high-level and low-level features. DS-TransUNet~\cite{lin2022ds} and Swin-UNet~\cite{cao2021swin} are transformer-based neural networks designed for 2D image segmentation, utilizing a U-shaped architecture built upon the Swin Transformer. They are considered "pure" transformer networks since they exclusively use transformer layers for feature extraction and processing, without depending on any pre-existing backbone architecture that has been pre-trained. The UNETR~\cite{hatamizadeh2022unetr} architecture uses a transformer-based encoder to embed input 3D patches, followed by a CNN-based decoder that produces the final 3D segmentation output. In contrast, the HiFormer~\cite{heidari2023hiformer} model integrates a CNN and a Transformer by leveraging multi-scale feature representations from the Swin Transformer block and a CNN-based encoder. Additionally, it employs a Double-Level Fusion module to merge global and local features.
TransNorm~\cite{azad2022transnorm} describes a technique that incorporates a Transformer module into the encoder and skip-connections of a standard U-Net. By using skip-connections, it merges features between the expanding and contracting paths. Meanwhile, TransDeepLab is based on DeepLab~\cite{chen2017deeplab} and utilizes a purely Transformer-based model. This approach implements a hierarchical Swin-Transformer with shifted windows to create the Atrous Spatial Pyramid Pooling module.
TFCNs~\cite{li2022tfcns} enhance FC-DenseNet by introducing the ResLinear-Transformer and Convolutional Linear Attention Block, increasing the extraction of features from CT images through the utilization of more latent information.
The FCT~\cite{tragakis2023fully} is a medical imaging model that combines CNNs and Transformers to process input in two stages, first extracting long-range semantic dependencies and then capturing hierarchical global attributes from the features. In hybrid systems, basic techniques for combining features are used that do not ensure consistency between various levels of scale. To address this issue, we introduce Dilated-Unet, a transformer-based U-shaped architecture that effectively expands the local areas of focus and incorporates a global context into hierarchical visual transformers. This helps to preserve the richness and consistency of features for the purpose of segmenting 2D medical images.


\section{Method}\label{sec:method}

In this section, we will provide an overview of the proposed Dilated-Unet architecture. As shown in Figure~\ref{fig:overal}, the design blends global contextual and local representative features from the DiNA block in the encoder to produce hierarchical feature representations. Section 3.1 will delve into the DiNA block, while Sections 3.2 and 3.3 will delve into the specifics of the encoder and decoder components of the Dilated-Unet, respectively.

\begin{figure*}[ht]
    \centering
    \includegraphics[width=0.9\textwidth]{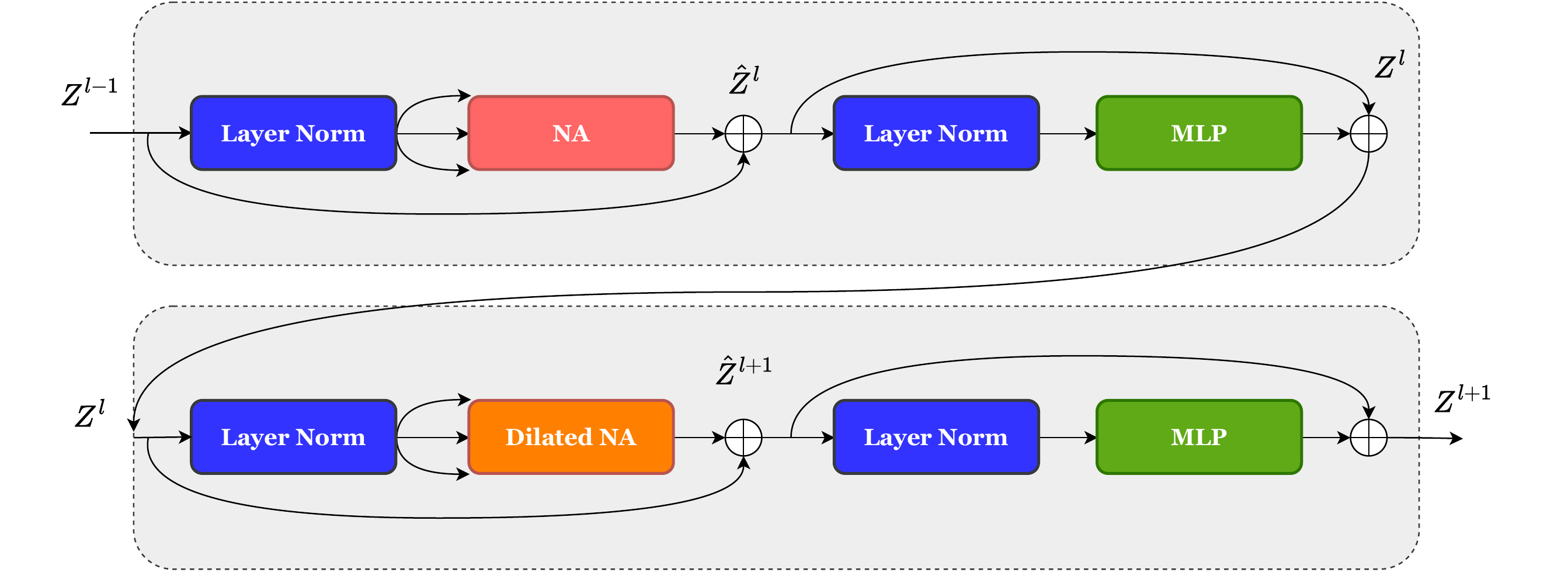}
    \vspace{1em}
    \caption{The overview of Dilated Transformer block, which is composed of NA, MLP, DiNa and Layer Norms.}
    \label{fig:Dilated}
    \vspace{2em}
\end{figure*}

\subsection{DiNA Transformer block}
The DiNA block, described in \cite{hassani2022dilated}, differs from the traditional multi-head self attention (MHSA) module and serves as a sparse, global operation that is most effective when paired with the NA module as a local-only operation. Figure~\ref{fig:Dilated} shows two consecutive DiNA transformer blocks, each consisting of LayerNorm (LN), Neighborhood Attention (NA), Dilated Neighborhood Attention (DiNA), a residual connection, and a 2-layer MLP with a GELU non-linearity. The transformer blocks utilize both the sliding-window Neighborhood Attention (NA) and Dilated Neighborhood Attention (DiNA) modules, which contribute to the flexible and efficient attention mechanism. As a result, the continuous application of DiNA Transformer blocks can be implemented:

\begin{align}
&\hat{z}^l=N A\left(L N\left(z^{l-1}\right)\right)+z^{l-1}, \\
&z^l=M L P\left(L N\left(\hat{z}^l\right)\right)+\hat{z}^l, \\
&\hat{z}^{l+1}=Dilated-N A\left(L N\left(z^l\right)\right)+z^l, \\
&z^{l+1}=\operatorname{MLP}\left(L N\left(\hat{z}^{l+1}\right)\right)+\hat{z}^{l+1},
\end{align}

where $\hat{z}^l$ and $z^l$ represent the outputs of the $NA$ module and the MLP module of the $l^{t h}$ block, respectively. Similar to the previous work \cite{hassani2022dilated}, Dilated neighborhood self-attention is computed as follows:

\begin{equation}
\operatorname{DiNA}_k^\delta(i)=\operatorname{softmax}\left(\frac{\mathbf{A}_i^{(k, \delta)}}{\sqrt{d_k}}\right) \mathbf{V}_i^{(k, \delta)},
\end{equation}

where DiNA is output for the $i$-th token neighborhood size $k$. ${V}_i^{(k, \delta)}$ and ${A}_i^{(k, \delta)}$ are values and weights of $\delta$-dilated neighborhood attention for the $i$-th token, respectively. Also, $\sqrt{d_k}$ is the scaling parameter, and $d_k$ is the key dimension.

\subsection{Encoder}
The encoder in this system processes tokenized inputs of dimension C and resolution of $\frac{H}{4} \times \frac{W}{4}$. These inputs are passed through two Dilated Transformer blocks for representation learning, where the feature dimension and resolution remain unchanged. However, the patch merging layer reduces the number of tokens by a factor of 2 through downsampling, and increases the feature dimension by a factor of 2. This process is repeated three times in the encoder. The patch merging layer divides the input patches into four parts, which are then concatenated together. This concatenation operation results in a downsampling of the feature resolution by a factor of 2 and an increase in the feature dimension by a factor of 4. To mitigate this increase, a linear layer is applied on the concatenated features to bring the feature dimension back to double the original dimension.

\begin{table*}[!ht]
    \centering
    \caption{Comparison results of the proposed method on the \textit{Synapse} dataset. \textcolor{blue}{Blue} indicates the best result, and  \textcolor{red}{red} indicates the second-best.}
    \vspace{0.5em}
    \resizebox{\textwidth}{!}{
        \begin{tabular}{l|cc|cccccccc}
            \toprule
            \textbf{Methods}                        & \textbf{DSC~$\uparrow$} & \textbf{HD~$\downarrow$} & \textbf{Aorta}          & \textbf{Gallbladder}    & \textbf{Kidney(L)}      & \textbf{Kidney(R)}      & \textbf{Liver}          & \textbf{Pancreas}       & \textbf{Spleen}         & \textbf{Stomach}        \\
            \midrule
            DARR \cite{fu2020domain}                & 69.77                   & -                        & 74.74                   & 53.77                   & 72.31                   & 73.24                   & 94.08                   & 54.18                   & 89.90                   & 45.96
            \\
            R50 U-Net \cite{chen2021transunet}      & 74.68                   & 36.87                    & 87.74                   & 63.66                   & 80.60                   & 78.19                   & 93.74                   & 56.90                   & 85.87                   & 74.16
            \\
            U-Net \cite{ronneberger2015u}           & 76.85                   & 39.70                    & 89.07  & \textcolor{red}{69.72}  & 77.77                   & 68.60                   & 93.43                   & 53.98                   & 86.67                   & 75.58
            \\
            R50 Att-UNet \cite{chen2021transunet}   & 75.57                   & 36.97                    & 55.92                   & 63.91                   & 79.20                   & 72.71                   & 93.56                   & 49.37                   & 87.19                   & 74.95
            \\
            Att-UNet \cite{oktay2018attention}  & 77.77                   & 36.02                    & \textcolor{blue}{89.55} & 68.88                   & 77.98                   & 71.11                   & 93.57                   & 58.04                   & 87.30                   & 75.75
            \\
            R50 ViT \cite{chen2021transunet}        & 71.29                   & 32.87                    & 73.73                   & 55.13                   & 75.80                   & 72.20                   & 91.51                   & 45.99                   & 81.99                   & 73.95
            \\
            TransUNet \cite{chen2021transunet}      & 77.48                   & 31.69                    & 87.23                   & 63.13                   & 81.87                   & 77.02                   & 94.08                   & 55.86                   & 85.08                   & 75.62
            \\
            Swin-Unet \cite{cao2021swin}            & 79.13                   & 21.55                    & 85.47                   & 66.53                   & 83.28                   & 79.61                   & 94.29                   & 56.58                   & 90.66                   & 76.60
            \\
            LeVit-Unet \cite{xu2021levit}           & 78.53                   & \textcolor{red}{16.84}                    & 78.53                   & 62.23                   & 84.61                   & 80.25                   & 93.11                   & 59.07                   & 88.86                   & 72.76
            \\
            MISSFormer \cite{huang2021missformer}   & \textcolor{red}{81.96}  & 18.20                    & 86.99                   & 68.65                   & 85.21                   & \textcolor{blue}{82.00}                   & 94.41                   & \textcolor{blue}{65.67} & \textcolor{red}{91.92}  & 80.81
            \\
            DeepLabv3+ (CNN) \cite{chen2017deeplab} & 77.63                   & 39.95                    & 88.04                   & 66.51                   & 82.76                   & 74.21                   & 91.23                   & 58.32                   & 87.43                   & 73.53
            \\
            HiFormer \cite{heidari2023hiformer}     & 80.39                   & \textcolor{blue}{14.70}  & 86.21                   & 65.69                   & \textcolor{red}{85.23}  & 79.77                   & \textcolor{red}{94.61}                   & 59.52                   & 90.99                   & \textcolor{red}{81.08}
            \\
            \hline
            \rowcolor{cyan!10}
            \textbf{Dilated-Unet}            & \textcolor{blue}{82.43} & 17.46                    & \textcolor{red}{89.16}                   & \textcolor{blue}{72.30} & \textcolor{blue}{86.08} & \textcolor{red}{81.40}  & \textcolor{blue}{94.98} & \textcolor{red}{65.12}  & \textcolor{blue}{91.94} & \textcolor{blue} {81.19}
            \\
            \bottomrule
        \end{tabular}
    }\label{tab:performance_comparison}
\end{table*}

\begin{figure*}[ht]
    \centering
    \includegraphics[width=0.8\textwidth]{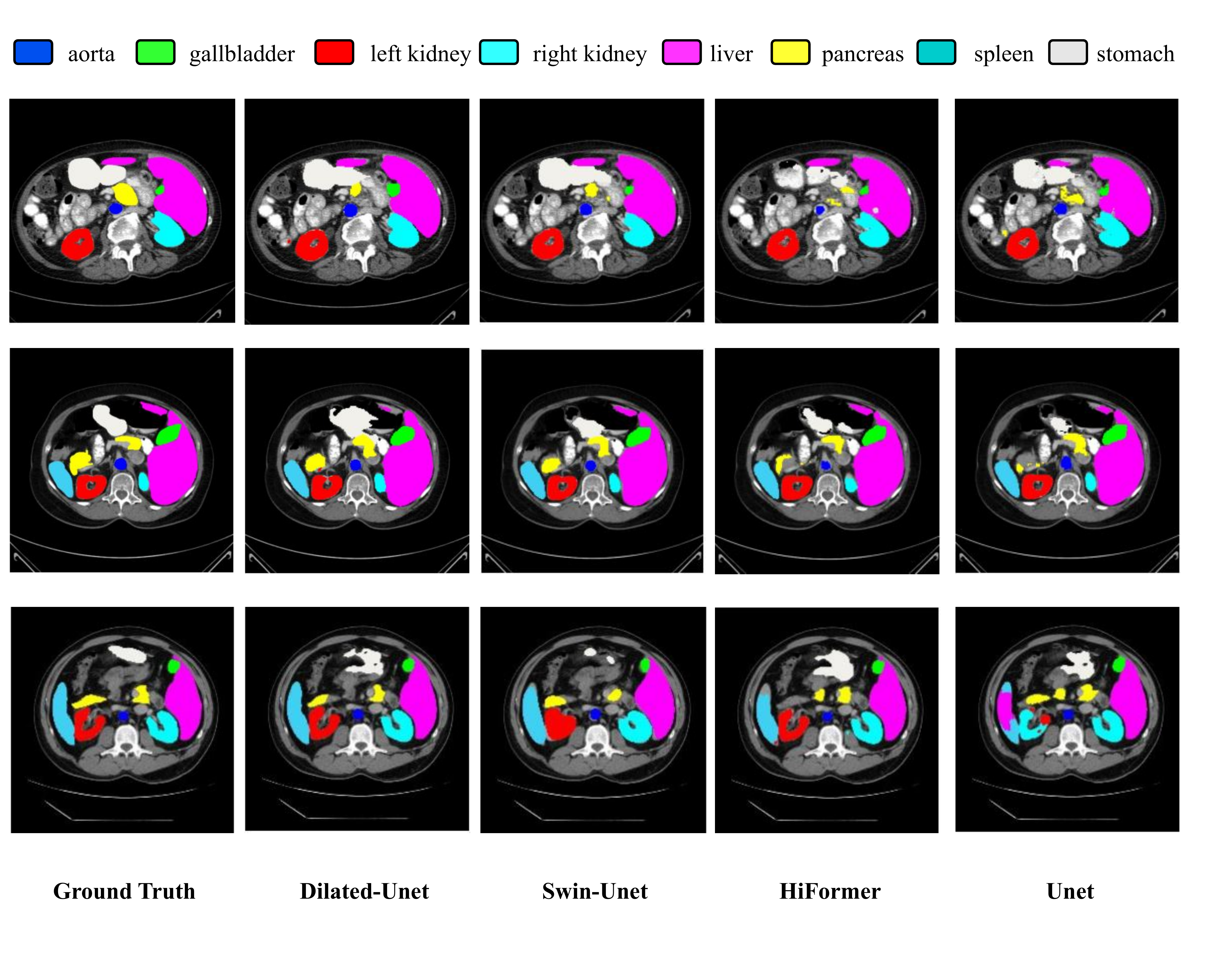}
    \caption{Comparative segmentation results on the \textit{Synapse} dataset.}
    \label{fig:synapsevisualization}
\end{figure*}

\subsection{Bottleneck}

The Transformer architecture is known to be quite deep and complex, making it difficult for the model to converge during training~\cite{touvron2021going}. To overcome this issue, only two successive Dilated Transformer blocks are utilized to construct the bottleneck. The purpose of the bottleneck is to learn the deep feature representation, which can be used for various applications. In order to maintain the quality of the representation, the feature dimension and resolution are kept unchanged within the bottleneck. This allows the model to focus on learning the underlying patterns in the data, rather than being influenced by changes in feature dimension or resolution. By utilizing this approach, the model can achieve robust results while avoiding the convergence issues associated with deep Transformer models.

\subsection{Decoder}
The symmetrical decoder is designed to match the encoder and is based on the Dilated Transformer block. In order to upscale the extracted deep features, the patch expanding layer is utilized in the decoder instead of the patch merging layer used in the encoder. This layer transforms the feature maps from adjacent dimensions into a higher resolution feature map through $2 \times$up-sampling, which reduces the feature dimension to half of its original size. 
As an illustration, consider the first patch expanding layer. Before the upscaling process, a linear layer is applied to the input features to double their dimension, resulting in $\left(\frac{W}{32} \times \frac{H}{32} \times 8 C\right)$ which is $2 \times$ the original dimension $\left(\frac{W}{32} \times \frac{H}{32} \times 16 C\right)$. Then, a rearrange operation is performed to expand the resolution of the input features to $2 \times$ their original resolution, while also reducing the feature dimension to one quarter of the input dimension, $\left(\frac{W}{32} \times \frac{H}{32} \times 16 C \rightarrow \frac{W}{16} \times \frac{H}{16} \times 4 C\right)$. The impact of using the patch expanding layer in the upscaling process will be analyzed in section 4.5.

\subsection{Skip connection}

The model follows the approach of the U-Net~\cite{ronneberger2015u} by utilizing skip connections to merge the multi-scale features from the encoder with the up-sampled features. This helps to prevent the loss of spatial information caused by down-sampling. To achieve this, the shallow and deep features are concatenated and then passed through a linear layer, which maintains the dimension of the concatenated features to be the same as the up-sampled features. The impact of the number of skip connections on the performance of the model will be thoroughly analyzed in section 4.5.

\section{Experimental Result}
The proposed method for medical image segmentation was validated through experiments on two tasks. Initially in this section, the datasets used for evaluation are introduced, and the metrics and experimental settings are explained. The obtained outcomes are examined through the presentation of quantitative and visualization results for each dataset. Impartiality is ensured by comparing our proposed model with CNN-based, transformer-based, and combination models.

\subsection{Dataset}
\textbf{Synapse multi-organ segmentation.} The synapse multi-organ segmentation dataset~\cite{synapse2015ct} includes 30 abdominal CT scans consisting of a total of 3779 axial abdominal clinical CT images. Each CT volume encompasses varying dimensions with 512 x 512 x 85 to 512 x 512 x 198 pixels and field of views ranging from approximately 280 x 280 x 280 mm3 to 500 x 500 x 650 mm3. The spatial resolution of each voxel ranges from 0.54 x 0.54 mm2 to 0.98 x 0.98 mm2, and the thickness of each slice varies from 2.5 mm to 5.0 mm. In each scan, 13 organs were identified and labeled by interpreters, including the gallbladder, the spleen, the liver, the left kidney, the right kidney, the esophagus, the aorta, the inferior vena cava, the stomach, the right adrenal gland, and the left adrenal gland, the splenic vein, the portal vein,  the pancreas. 

\textbf{Skin Lesion Segmentation.} In 2018, the ISIC Foundation created a comprehensive dataset of dermoscopy images \cite{codella2019skin}. The dataset comprises over 10,000 images representing 7 types of skin diseases including seborrheic keratosis, nevi, melanoma, BCC, actinic keratosis, Bowen’s disease, dermatofibromas, and vascular lesions, along with their corresponding annotated ground truths.

\begin{figure*}[ht]
    \centering
    \includegraphics[width=0.8\textwidth]{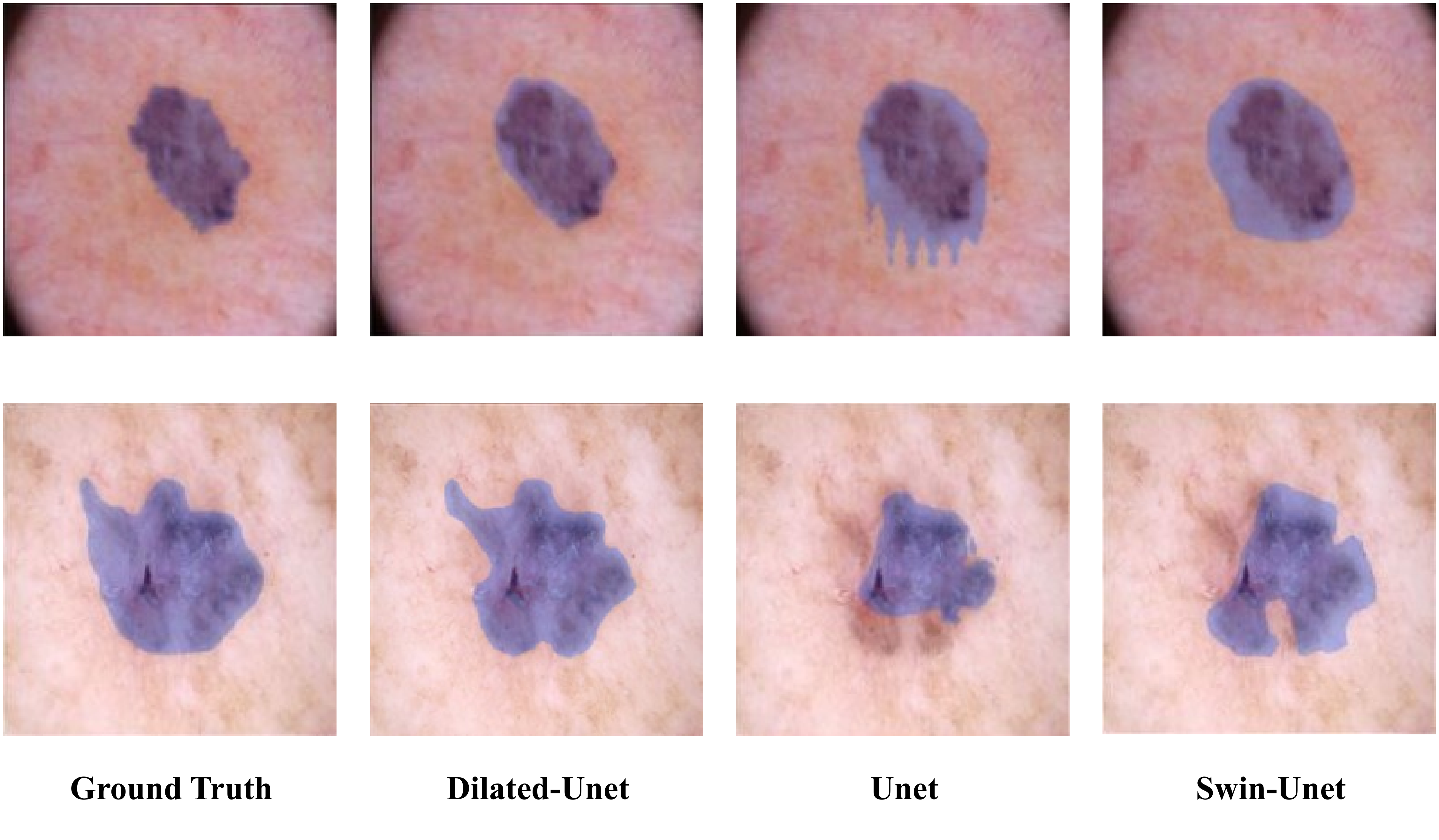}
    \caption{Comparative segmentation results on the \textit{Synapse} dataset.}
    \label{fig:ISICvisualization}
\end{figure*}

\begin{table}[h]
    \centering
    \caption{Performance comparison of the proposed method against the SOTA approaches on the \textit{ISIC2018} skin lesion segmentation task.} \label{tab:quantitative_ISIC}
    \resizebox{\columnwidth}{!}{
        \begin{tabular}{l || c c c c }
            \hline
            \textbf{Methods}                    & \textbf{DSC}    & \textbf{SE}     & \textbf{SP}     & \textbf{ACC}    \\
            \hline
            U-Net \cite{ronneberger2015u}       & 0.8545          & 0.8800          & 0.9697          & 0.9404          \\
            Att U-Net \cite{oktay2018attention} & 0.8566          & 0.8674          & \textbf{0.9863} & 0.9376          \\
            TransUNet \cite{chen2021transunet}  & 0.8499          & 0.8578          & 0.9653          & 0.9452          \\
            MedT \cite{valanarasu2021medical}   & 0.8389          & 0.8252          & 0.9637          & 0.9358          \\
            FAT-Net \cite{wu2022fat}            & 0.8903          & 0.9100 & 0.9699          & 0.9578          \\
            TMU-Net \cite{azad2022contextual}   & 0.9059          & 0.9038          & 0.9746          & 0.9603          \\
            Swin\,U-Net \cite{cao2021swin}      & 0.8946          & 0.9056          & 0.9798          & 0.9645          \\
            TransNorm \cite{azad2022transnorm}  & 0.8951          & 0.8750          & 0.9790          & 0.9580          \\
            \hline
            \rowcolor{cyan!10}
            \textbf{Dilated-Unet} & \textbf{0.9147} & \textbf{0.9129} & 0.9853 & \textbf{0.9647}          \\ \hline
        \end{tabular}
    }
\end{table}

\subsection{Evaluation Metrics}
The performance of Synapse multi-organ segmentation was assessed by computing the average Hausdorff Distance (HD) and the average Dice Similarity Coefficient (DSC). In order to evaluate the results of skin lesion segmentation, we used several widely recognized metrics, which include F1-Score, sensitivity, accuracy (ACC), and  specificity. The equations for each metric are provided below along with a brief explanation. The symbol TP denotes True Positive, which represents a sample that has been accurately predicted as a positive sample. FP corresponds to False Positive, which indicates that the model predicted a negative sample as a positive sample, when the true label was negative. TN represents True Negative, which refers to a sample with a negative label that has been correctly identified as a negative sample. FN denotes False Negative, which indicates an incorrect classification of a true positive sample.\\
\textbf{Specificity} represented as True Negative (TN) rate:
\begin{equation}
\text { Specificity }=\frac{T N}{T N+F P}
\end{equation}\\
\textbf{Sensitivity} represented as True Positive (TP) rate: 
\begin{equation}
\text { Sensitivity }=\frac{T P}{T P+F N}
\end{equation}\\
\textbf{Accuracy} representation of the correct classification rate:
\begin{equation}
\text {ACC}=\frac{T P+T N}{T P+T N+F P+F N}
\end{equation}\\
\textbf{DSC} calculates the similarity between the predicted segmentation and the ground truth segmentation:
\begin{equation}
\text {DSC} =\frac{2T P}{2T P+T N+F P+F N}
\end{equation}

\subsection{Implementation Details}
In this research, all experiments were conducted using the Python programming language version 3.8 and the PyTorch deep learning library version 1.8.0. To enhance the diversity of the training data, augmentation techniques such as random flipping and rotations were applied to all training samples. The images used for training were resized to 224 x 224, and the training was performed on Nvidia 3090 GPUs with 24GB of memory.
The models were trained using the Adam optimization algorithm with a learning rate of 1e-5 and weight decay of 1e-4. The objective function used during training was a combination of cross entropy and Dice loss. The training epochs were set to 350 for the Synapse dataset and 400 for the ISIC dataset.
To improve the performance of the CNNs and Transformers, their parameters were initialized using pre-trained weights from the ImageNet dataset.

\section{Evaluation Results}

\subsection{Results of Synapse Multi-Organ Segmentation}
The comparison of the proposed method with previous state-of-the-art techniques in terms of average Dice Similarity Coefficient (DSC) and average Hausdorff Distance (HD) on eight abdominal organs is shown in Table? The results presented indicate that our proposed model significantly outperforms pure CNN models. Our model demonstrates superior learning ability compared to other transformer-based models as evidenced by a decrease of 14.23 and 4.09 in average HD and an increase of 4.95\% and 3.3\% in Dice score andcompared to TransUnet~\cite{chen2021transunet} and Swin-Unet~\cite{cao2021swin}, respectively. Furthermore, when compared to hybrid CNN-transformer approaches such as MISSFormer~\cite{huang2021missformer}, our method improves the average DSC and Hausdorff distance by 0.47\% and 0.74 mm, respectively, and records a 2.04\% increase in Dice score compared to HiFormer~\cite{heidari2023hiformer}. Specifically, our model outperforms previous literature in the segmentation of most organs, particularly the liver, spleen, left kidney, stomach, and gallbladder, achieving the best results. For the remaining organs, our model obtains the second-best results. Additionally, for visual evaluation, we have presented a representative qualitative example in Figure~\ref{fig:synapsevisualization} which demonstrates the superior performance of the proposed method, as it more closely resembles the ground truth and accurately reflects the actual situation.

\subsection{Results of Skin Lesion Segmentation}
The comparison of the ISIC 2018 skin lesion segmentation task against top-performing methods is presented in Table~\ref{tab:quantitative_ISIC}. As with the synapse dataset, our method surpasses both Transformers and CNN-based approaches in terms of most of the evaluation metrics. Additionally, a visual comparison of the skin lesion segmentation results is presented in Figure~\ref{fig:ISICvisualization}, which demonstrates that the proposed method is capable of identifying finer structures and creating more precise contours. Despite the presence of overlap between the background and the skin lesion class, the method still yields highly accurate segmentation results. Our approach outperforms models such as the Swin-UNet~\cite{cao2021swin} in terms of boundary precision and robustness to noise.

\begin{table*}[h]
\centering
\caption{Impact of the number of skip-connection in Dilated-UNet.}
\footnotesize
\resizebox{\textwidth}{!}{
\begin{tabular}{c|cc|cccccccc}
\hline
Number of skip-connection &  DSC$\uparrow$ &HD$\downarrow$ & Aorta& Gallbladder& Kidney(L)& Kidney(R)& Liver& Pancreas& Spleen& Stomach   \\
\hline
0 & 72.62  & 28.6  & 82.99  & 62.14  & 73.02  & 66.94  & 93.36  & 55.96  & 88.34  & 75.24  \\
1 & 78.44  & 26.47  & 86.81  & 69.34  & 76.90  & 70.31  & 94.40  & 63.25  & 90.26  & 79.01  \\
2 & 80.81  & 21.49  & 88.72  & 71.08  & 83.04  & 77.33  & 94.73  & 62.05  & 90.70  & \textbf{81.62}  \\
3 & \textbf{82.43} & \textbf{17.46} & \textbf{89.16} & \textbf{72.30} & \textbf{86.08} & \textbf{81.40}  & \textbf{94.98} & \textbf{65.12}  & \textbf{91.94} & 81.19\\
\hline
\end{tabular}
}
\label{skip_conection}
\end{table*}

\begin{table*}[h]
\centering
\caption{Effect of input size on the performance of the Dilated-UNet.}
\footnotesize
\resizebox{\textwidth}{!}{
\begin{tabular}{c|cc|cccccccc}
\hline
Input size &  DSC$\uparrow$ &HD$\downarrow$ & Aorta& Gallbladder& Kidney(L)& Kidney(R)& Liver& Pancreas& Spleen& Stomach   \\
\hline
224 &  82.43 & 17.46 & 89.16 & 72.30 & 86.08 & 81.40  & 94.98 & 65.12 & \textbf{91.94} & \textbf{81.19}  \\
384 & \textbf{84.42} & \textbf{15.28} & \textbf{90.76} & \textbf{79.35} & \textbf{87.44} & \textbf{84.66}  & \textbf{95.41} & \textbf{70.27}  & 91.42 & 79.88\\
\hline
\end{tabular}
}
\label{input_size}
\end{table*}

\subsection{Ablation study}
In order to understand the impact of various factors on the performance of the proposed Dilated-UNet architecture, we carried out an ablation study using the Synapse dataset. The study focused on examining the effect of the number of skip connections, the size of the inputs, and the size of the model, which are discussed in detail below.

\textbf{Impact of the number of skip connections:}
We studied the effect of incorporating skip-connections into the Dilated-UNet model. The findings are displayed in Table~\ref{skip_conection}. It's important to note that the "1-skip" setting refers to using one skip-connection at the 1/4 resolution level, while "2-skip, and "3-skip"  refer to incorporating skip-connections at resolutions 1/8, and 1/16, respectively. Our results show that incorporating more skip-connections leads to improved performance. Additionally, smaller organs such as the aorta, gallbladder, and kidneys showed a more significant improvement than larger organs like the liver, spleen, and stomach. As a result, in order to enhance the model's robustness, the number of skip-connections was set to 3 in this study.

\begin{table}[h]
    \centering
    \caption{Comparison of model parameters.}
    \vspace{-0.5em}
    \label{table:model parameters}
    \resizebox{0.9\columnwidth}{!}{%
    \begin{tabular}{cccc}
    \hline
     Model & \multicolumn{1}{l}{\# Params (M)} & \multicolumn{1}{l}{{DSC $\uparrow$}} & \multicolumn{1}{l}{HD $\downarrow$} \\ \hline
    TransUnet   & 105.28 & 77.48 &  31.69      \\
    Swin-Unet   & 27.17 & 79.13 & 21.55      \\
    LeVit-Unet   & 52.17 & 78.53 & 16.84     \\
    DeepLabv3+ (CNN)  & 59.50 & 77.63 &  39.95      \\
    \midrule
    \textbf{Dilated-Unet-tiny}  & \textbf{25.15} & \textbf{82.43} &  \textbf{17.46}      \\
    \textbf{Dilated-Unet-small}  & \textbf{28.51} & \textbf{82.59} & \textbf{13.42}      \\
    \textbf{Dilated-Unet-large}  & \textbf{35.73} & \textbf{82.91} &  \textbf{17.74}     \\ \hline
    \end{tabular} %
    }
    \vspace{1em}
\end{table}

\textbf{Impact of input size:}
Table~\ref{input_size} displays the results of ablation study on the introduced Dilated-Unet with input resolutions of 224 x 224 and 384 x 384. When the input size is increased from 224 x 224 to 384 x 384 while the patch size remains at 8, the input token sequence of the Transformer gets longer, which improves the model's segmentation capabilities. However, this improvement in segmentation accuracy comes at a cost, as the computational demand of the entire network has considerably increased. In order to maintain the algorithm's efficiency, the experiments in this study opted for a 224 x 224 resolution scale as the input, which strikes a balance between computational complexity and accuracy.

\textbf{Impact of model scale:}
We examine the impact of deepening the network on model performance, similar to what was done in~\cite{cao2021swin}. The results in Table~\ref{table:model parameters} indicate that expanding the size of the model does not significantly enhance its performance, but raises the overall computational cost of the model. Given the trade-off between accuracy and speed, we choose to utilize the tiny-based architecture for medical image segmentation.

\section{Conclusion}

{\small
\bibliographystyle{ieee_fullname}
\bibliography{egbib}
}

\end{document}